\documentclass{article}

\usepackage[english]{babel}

\usepackage[letterpaper,top=2cm,bottom=2cm,left=3cm,right=3cm,marginparwidth=1.75cm]{geometry}

\usepackage{amsmath}
\usepackage{amsfonts}
\usepackage{graphicx}
\usepackage[colorlinks=true, allcolors=blue]{hyperref}
\usepackage{adjustbox}
\usepackage{subcaption}
\usepackage{threeparttable}
\usepackage{booktabs}

\usepackage{amssymb}
\usepackage{algorithm}
\usepackage{algpseudocode}
\usepackage{float}
\usepackage{array}
\usepackage{graphicx}
\usepackage{pgfplots}

\pgfplotsset{compat=1.18}
\usepgfplotslibrary{colormaps}

\title{Optimization with Dynamic Constraint Learning\\ (DCL)}
\author{Ezgi Öztekin\thanks{Department of Mathematics, Gebze Technical University}, Figen Öztoprak\thanks{Department of Industrial Engineering, Gebze Technical University} and Ş. İlker Birbil\thanks{Amsterdam Business School, University of Amsterdam}}

\begin{document}
\maketitle

\begin{abstract}
We propose Dynamic Constraint Learning (DCL), a data-driven framework for constrained optimization when constraint functions are unknown and cannot be queried during optimization. At each iteration, the method learns a local surrogate from nearby data and solves a subproblem within a data-supported trust region. Compared with offline global constraint learning, the approach uses local surrogates that adapt to the data distribution during optimization and can achieve solution quality comparable to that of global models while using simpler local models and smaller optimization subproblems.  We demonstrate the performance of DCL on a synthetic test problem and two case studies from the literature.
\end{abstract}


\section{Introduction}
We consider constrained optimization problems in which the analytic form of at least one constraint function is unknown and cannot be queried during optimization. To this end, we assume a \emph{learning} setup: We have labeled data at hand with labels corresponding to noisy evaluations of the unknown constraint function with uncertainties, and we aim at satisfying the constraint for the population via a prediction model.  

A prominent example is radiotherapy treatment planning \cite{maragno2024}: the optimization model must limit damage to healthy tissue, yet the dose-response relationship depends on patient-specific biological factors that cannot be expressed analytically.  The feasibility requirement can only be \emph{predicted} from data.  Similar situations arise in diet planning (palatability constraints with no analytical score) \cite{opticl}, structural engineering (failure criteria estimated from load-testing data), and operations management (service-level constraints inferred from historical demand records).


The predominant approach in the literature follows a \emph{two-phase} pipeline: first, fit a global surrogate for the unknown constraint using all available data; then, embed that surrogate into an optimization model and solve it once \cite{fajemisin2024,bergman2022,opticl}.  This estimate-then-optimize paradigm is natural and practically effective.  However, it has a structural limitation: the surrogate is trained to be accurate \emph{everywhere} over the domain, yet optimization ultimately cares only about accuracy \emph{near the optimal solution}.  Achieving global accuracy often requires complex model classes---deep neural networks, high-degree polynomials---which in turn produce
hard optimization subproblems \cite{yang2022}.


This work proposes an alternative approach that we call \emph{Dynamic Constraint Learning (DCL)}.  DCL is a framework that learns local models of the unknown constraint function in the process of optimization.  Local models are learned at each (outer) iteration of the optimization process using subsets of the available data.  In some neighborhood of each iterate, the model class and the data subset is determined in a dynamical manner. 
These local models are then used to compute steps in the variable space.  We conjecture that both the learning and optimization components can benefit from local surrogate models.  In particular, locally learned models could be less complex and use less data points, which would create simpler optimization subproblems.  Besides, they could provide better prediction models in the neighborhoods important for optimization.  Moreover, since these models would potentially be simpler, they could be \emph{interpretable} at the iterates of the optimization run including the final solution. 

From a stochastic optimization perspective, global constraint learning resembles Sample Average Approximation (SAA): it is an \emph{offline} approach that produces a fixed model of the unknown constraint for the optimization phase.  On the contrary, DCL is an \emph{online} approach that employs subsets of the available data at different iterations of the optimization process.  The models produced by DCL at different iterations would contain stochastic error.  In that sense, DCL can be seen as a \emph{stochastic approximation (SA)} method. 

Very complex machine learning models such as neural networks could be needed for global surrogate fitting as the model is desired to be successful over the whole domain of optimization.  However, complex models lead to hard optimization problems \cite{yang2022}.  Indeed, constrained optimization methods generally do not rely on global analytical forms of complex constraint functions.  At each iteration, these methods build a locally approximate model of the problem.  That general scheme fits to model-based derivative free optimization (DFO) as well as derivative-based methods such as sequential quadratic programming.  This observation motivates building local surrogates rather than global ones in particular when there is no guarantee for computing a global solution. 


We note finally that although we focus on \emph{learned constraints}, the framework also covers the case where the objective function is estimated via a prediction model. This is possible because any objective function can be reformulated as a level-set constraint by introducing an auxiliary variable.


\subsection{Related work}
\label{sec:related}

\begin{figure}
\centering
\includegraphics[width=0.75\textwidth]{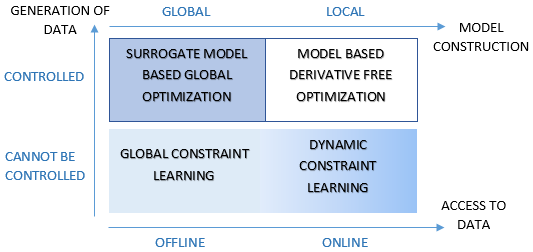}
\caption{Dynamic constraint learning as a model based optimization approach.  There is no boundary between global constraint learning and DCL since the \emph{degree of locality} of DCL models depends on the distribution of data.}
\label{fig:dcl}
\end{figure}

Surrogate models are used in optimization across several distinct settings that differ in two fundamental ways: whether the surrogate is an \emph{interpolation} model (fitted to data with non-noisy outputs) or a \emph{prediction} model (fitted to observational data with uncertainty), and whether new labeled data can be \emph{generated on demand} or must be drawn from a fixed database.  DCL operates in the prediction/fixed-data quadrant of this space, illustrated in Figure \ref{fig:dcl}.  We now review the four bodies of work closest to DCL.


\paragraph{Surrogate-model-based optimization.}  Learning an algebraic model from data obtained via simulations or experimentation is a central issue in simulation optimization.  In this setup, data generation can be controlled as the simulation (or the physical experiment) can be conducted at desired points.  Surrogate-model-based optimization techniques generally start by fitting a global surrogate model.  Since the final purpose of the generated model is optimization, functional forms appropriate for integration to optimization models are targeted \cite{cozad2014}.  The optimization phase can rely on local or global methods. The global surrogate is refined via \emph{infilling} until a satisfactory solution is found; the infill criteria may be motivated by exploration or exploitation \cite{forrester2008}.  A well-known surrogate-model-based optimization technique is \emph{Bayesian Optimization}. 

\paragraph{Constrained model-based DFO.} Model-based DFO methods rely on the idea of step computation based on local models of the problem.  The local models are generally produced via interpolation or regression, and special care is given to the geometric conditions to determine new evaluation points \cite{conn2009}.  As in global surrogate-based techniques, data can be generated at desired points; however, the design of model-based DFO methods rely on local models constructed at each iteration rather than a global model.  These local models are generally convex quadratic functions; there is recent work though that employ non-convex local models (such as RBF interpolation) within a model-based DFO framework \cite{wild2011}.

The design of DCL is largely inspired by model-based DFO techniques.  However, there are big differences due to the nature of the problem.  In DCL, the choice of the model complexity depends on the available data since it is not possible to generate data at desired points.  Due to the same reason, there is no clear way to use a step computation trust region.  For DCL, the trust region --a region in which the local model is assumed to be valid-- is defined via the data subset used for generating that model.


\paragraph{Global constraint learning.} As mentioned before, the idea of using prediction models in an embedded way to an optimization model has recently been investigated in the literature \cite{bergman2022, opticl}.  In existing work, the optimization phase follows the learning phase, and the optimization model constructed with the learned global surrogate is solved once.  The process of optimization with constraint learning has been formalized in \cite{fajemisin2024}, with a review of how the steps of this process are handled in different work. Computational tools has been developed that significantly facilitate practical implementations of the idea \cite{opticl,bergman2022,ceccon2022}.  An interesting extension of constraint learning to data-driven stochastic optimization is \emph{chance constraint learning} \cite{alcantara2023}, where a conditional quartile estimation is embedded into the optimization model rather than a point estimation.  

\paragraph{End-to-end constrained optimization learning.} Another recent idea relevant to our work is the so-called \emph{end-to-end constrained learning (E2ECO)}.  Similar to the setup for constraint learning, E2ECO might consider the case when some parameters of the optimization model are unknown and the learning problem can cover the prediction of these parameters.  The distinctive idea is to use the solution of the optimization model to compute the loss function, employing the solution quality of the optimization model for training rather than the prediction error \cite{kotary,tang2024pyepo}. E2ECO requires differentiability of the optimal solution map, and the idea has been investigated mostly on linear and quadratic programs. 


\subsection{Motivations for dynamic learning}

Global constraint learning faces a fundamental mismatch: it trains a model to be accurate over the entire domain, but optimization ultimately visits only a low-dimensional path through that domain.  When data is unevenly distributed---as is common in practice---a global model must either be highly complex to fit accurately across the data support, or accept poor accuracy in the regions that happen to matter most for optimization.  Either outcome is costly: complex models produce hard optimization subproblems \cite{yang2022}, while poor local accuracy degrades solution quality.

\begin{figure}[h!]
\centering
\includegraphics[width=\textwidth]{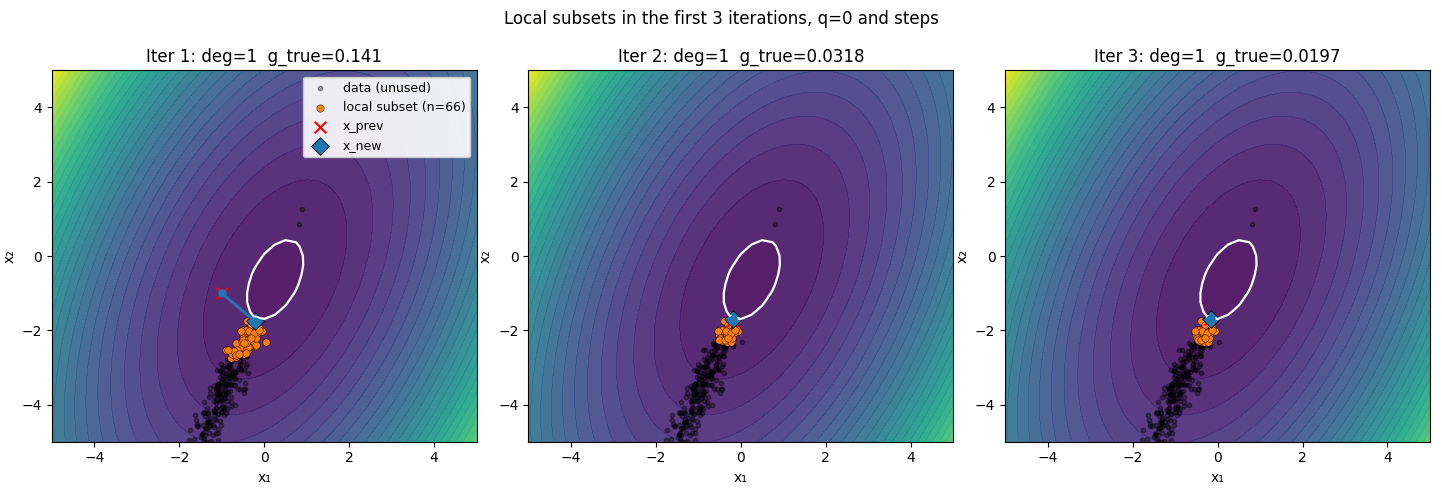}
\caption{First three DCL iterations for the quadratic constraint. Orange: used local subset;
gray: unused data; white curve: $q(x_1,x_2)=0$; red $\times$: previous iterate; blue $\diamond$:
new iterate. Each panel title shows the selected degree $d_k$ and the true constraint value
at the new point ($g_{\text{true}}$).}
\label{fig:first3-quadratic}
\end{figure}

DCL addresses this mismatch directly.  By restricting each surrogate to a neighborhood of the current iterate, the method (i)~needs only a fraction of the available data, (ii)~can employ a simpler hypothesis class adequate for the local geometry, and (iii)~produces a smaller optimization subproblem at each iteration.  The price is that any single local model may be inaccurate far from its training center---but that is precisely the region it does not need to represent.
\paragraph{An illustrative example.} To motivate the idea, let us first illustrate it on a simple example.  In this example, the true underlying constraint is a quadratic inequality constraint.  However, we do not know the analytic form of the constraint function and we also do not have any evaluation routines that return the value of the constraint function at a given point.  We only have a set of given data points $D$ which contain noisy evaluations of the constraint function at some solution points.  In Figure~\ref{fig:first3-quadratic}, we can see the true feasible region (boundary and interior of the white ellipse), and the set of data points at hand (all black and orange dots).  Let $q$ be the unknown underlying quadratic constraint function, i.e., the true constraint of the problem is
\[
g(x)=q(x_1,x_2)\le 0.
\]
At each iteration $k$, DCL constructs a local surrogate $\hat g_k$ of $g$ around the current iterate $x_k$ and then takes a step by using this local model.  The local models are trained by using a subset of the data around $x_k$.  In particular, distances from $x_k$ to all samples are computed and a nearby subset $D_k$ is selected (orange points in the figure).  Since we cannot generate additional data points, we try to achieve the best model available around $x_k$.  Each model construction process contains loops that fit models of increasing complexity and use datasets of increasing size, until no improvement in test RMSE can be achieved.  Once the local surrogate is obtained, a step is computed within the convex hull of the data points used in training the surrogate by solving  
 \[
    \min f(x)\quad\text{s.t.}\quad \hat g_k(x)\le 0,\qquad
    (x_1,x_2)\in \operatorname{conv}\{(x_1,x_2)_j:\, j\in D_k\},
  \]
which produces $x_{k+1}$.  

In this example, the local surrogate training process tries polynomial regression models with increasing degree until no improvement in test error is achieved.  Figure~\ref{fig:first3-quadratic} illustrates the first three iterations.
\begin{itemize}
  \item[] \textbf{Iteration 1.} Degree $d_1=1$; the selected local subset $D_1$ contains 66 nearby points. The step moves toward the
  feasibility boundary, with $g_{\text{true}}\approx 0.141>0$.
  \item[] \textbf{Iteration 2.} Degree $d_2=1$; the selected local subset $D_2$ contains 30 nearby points. The subset shifts toward the boundary and the step tracks it more closely, reaching $g_{\text{true}}\approx 0.0318$.
  \item[] \textbf{Iteration 3.} Degree $d_3=1$; the selected local subset $D_3$ contains 30 nearby points. The local fit learned from $D_3$ keeps the step within the convex hull and closer to the boundary, achieving
  $g_{\text{true}}\approx 0.0197$.
\end{itemize}

After three iterations, DCL is very close to the true solution of the problem.  For this simple instance and for the given data distribution, it figures out that it is enough to use local linear models of the constraint.  Overall, it uses a small portion of the whole dataset (the orange dots) and simple models to approach to a good solution.  


\bigskip

\paragraph{Properties of DCL.} We highlight four properties that distinguish DCL from the related approaches discussed in Section \ref{sec:related}.

\begin{itemize}
  \item \textbf{Local accuracy over global expressiveness.}  DCL conjectures that better solutions can be computed more efficiently than with global constraint learning, because local models achieve higher accuracy near each iterate with lower complexity. If the data distribution does not support reliable local fits anywhere, DCL gracefully reduces to global constraint learning.

  \item \textbf{Coupled but not end-to-end.}  DCL does not use optimization outcomes to train its surrogates, as E2ECO does.  Nevertheless, learning and optimization are tightly coupled: the model learning procedure targets accuracy around the iterates along the optimization path rather than average accuracy over the dataset; the optimization subproblems use these learned local models to compute the iterates; and the optimizer is discouraged to move into regions where data is too sparse to support a reliable model. Crucially, DCL does not require differentiability of the optimization model.

  \item \textbf{Interpretability along the optimization path.}  The sequence of local models generated by DCL constitutes a transparent account of how the solution was reached: each model describes the constraint geometry in the neighborhood of one iterate. The ensemble of local models can itself serve as a surrogate for use in downstream optimization. End-to-end methods provide no analogous explanation.

  \item \textbf{Online data regimes.}  DCL is well suited to settings where data arrives incrementally, such as a control system that receives periodic sensor measurements.  Because there is no stable global model to retrain, DCL can continue optimizing as new data are incorporated, with model generation and optimization proceeding in tandem.
\end{itemize}

\paragraph{Scenarios where DCL is most advantageous.} Two data regimes are particularly favorable.  First, when the dataset is large: abundant data around each iterate supports accurate local models, and DCL recovers the efficiency of model-based DFO without the cost of fitting a large global surrogate.  Second, when the data is spatially clustered: a global model fitted to clustered data may perform poorly in the optimization-relevant region, whereas a local model centered at the current iterate can be highly accurate where it matters.

\section{Dynamic Constraint Learning}

Dynamic constraint learning aims at solving the optimization problem
\begin{equation*}
\label{eq:dcl}
\begin{aligned}
\min_{x\in X} & \quad f(x) \\
\text{s.t.} & \quad c(x)\le 0,\\
& \quad g(x)\le 0.
\end{aligned}
\end{equation*}
Here, $f:X\rightarrow \mathbb{R}$ and $c:X\rightarrow \mathbb{R}^p$ are respectively the objective and constraint functions of the problem with known analytical forms. The function $g$ represents the unknown constraint to be estimated via the available data in set $D$.  The decision variables $x$ can be either continuous or integer valued. 

\subsection{The general framework of DCL}

The general framework of DCL consists of two basic components: Model construction and subproblem solve.  Algorithm~\ref{alg:dcl-framework} gives a description of DCL. 

At each iteration \(x_k\), a subset $D_k\subseteq D$ of nearby data points are selected and a local prediction model \(\hat g_k\) of the unknown constraint is constructed.  This step contains the learning procedure: choice of the local data subset, choice of the model family, training, and validation.  The details of local model learning are explained in Section \ref{subsec:model}.  If the best local model that can be obtained at \(x_k\) does not satisfy the required accuracy threshold, then the algorithm terminates.  Lack of finding a good fit (i.e., small test error) while gradually increasing the complexity of the learned model could occur when there is not enough data.  The solution quality of DCL clearly depends on the quality of the learned models.  This is why there is a threshold for the allowed test error for the local models. By the same reason, since the first local surrogate is constructed from the data available around \(x_0\), the initial point is chosen as a point around which data is available: a random point within the convex hull of $D$.

Once a local model $\hat g_k(x)$ is constructed, a step computation subproblem is solved to compute a trial point. 
\begin{equation}
\label{eq:subproblem}
    \min_{x\in X} \ f(x)
    \qquad
    \text{s.t.}\qquad
    c(x)\le 0,\quad
    \hat g_k(x)\le 0,\quad
    x\in \mathcal C_k
\end{equation}
The set $C_k$ is the \emph{trust region} for the learned model.  As detailed in Section \ref{subsec:tr}, it is defined based on the data $D_k$ used to construct the model $\hat g_k$.  In the present framework, the trust region $C_k$ is introduced precisely to keep the step inside a region where the local surrogate is expected to be reliable \cite{opticl,shi2024trust}.  No step acceptance mechanism is imposed to the algorithm at the outer level; however, the classical line search and trust region mechanisms could be employed at the inner level in solving subproblem \eqref{eq:subproblem}.  If the norm of the computed step becomes smaller than a prescribed threshold, then the algorithm terminates.


\begin{algorithm}[H]
\caption{Dynamic Constraint Learning (DCL)}
\label{alg:dcl-framework}
\begin{algorithmic}[1]
\Require Initial point $x_0\in X$, dataset $D=\{(z_i,y_i)\}_{i=1}^N$, tolerance $\varepsilon_{\rm step}>0$, accuracy threshold $\varepsilon_{\rm model}>0$, maximum number of iterations $K_{\max}$
\Ensure Approximate solution $x_k$

\State Set $k\gets 0$
\While{$k < K_{\max}$}
    \State Construct a local dataset $D_k$ around $x_k$
    \State Fit a local surrogate $\hat g_k$ of the unknown constraint using $D_k$
    \If{$\hat g_k$ does not satisfy the required accuracy threshold}
        \State \textbf{terminate}
    \EndIf

    \State Define a trust region $\mathcal C_k$ from the geometry of $D_k$
    \State Compute a trial point $x_k^{\rm trial}$ by solving the local subproblem \eqref{eq:subproblem}
    \If{$\|x_k^{\rm trial}-x_k\| < \varepsilon_{\rm step}$}
        \State Set $x_{k+1}\gets x_k^{\rm trial}$
        \State \textbf{terminate}
    \EndIf

    \State Set $x_{k+1}\gets x_k^{\rm trial}$
    \State $k\gets k+1$
\EndWhile
\end{algorithmic}
\end{algorithm}

\subsection{Matching available data with the model: Basic decisions and criteria}
\label{subsec:model}

A key question for DCL is how to match the amount of local data with the complexity of the model. If too few data points are used, the fitted model may be statistically unreliable. If too many data points are used, the model may lose locality and fail to represent the constraint accurately near the current iterate. DCL therefore adjusts both the local sample size and the model complexity dynamically.

\paragraph{Setting up \(D_k\).}
Given the current iterate \(x_k\), the local index set \(D_k\) is selected from the samples closest to \(x_k\). Let
\[
\delta_i^k := \|z_i-x_k\|, \qquad i=1,\dots,N,
\]
and let the sample indices be ordered so that
\[
\delta_{\pi_k(1)}^k \le \delta_{\pi_k(2)}^k \le \cdots \le \delta_{\pi_k(N)}^k.
\]
For a candidate sample size \(N_k^t\), define the candidate index set
\[
D_k^t := \{\pi_k(1),\dots,\pi_k(N_k^t)\}.
\]
Thus, \(D_k^t\) contains the indices of the \(N_k^t\) data points closest to \(x_k\).

Let \(d_k\) denote the degree of the local prediction model. For each candidate index set \(D_k^t\), prediction models of increasing degree are learned and evaluated by computing a validation error.  Once \emph{the best-fit model} is computed with dataset \(D_k^t\), a larger sample size $N_k^{t+1}>N_k^t$ is determined, and the model degree selection procedure is repeated with $D_k^{t+1}\supset D_k^t$.  The corresponding data subset and local model selection procedure is summarized in Algorithm~\ref{alg:dcl-local-model}.  The overall procedure terminates when no better test error can be achieved via a larger data subset.

\begin{algorithm}[H]
\caption{Local data selection and surrogate construction at iterate $x_k$}
\label{alg:dcl-local-model}
\begin{algorithmic}[1]
\Require Current iterate $x_k$, ordered sample indices $\pi_k(1),\ldots,\pi_k(N)$, initial sample size $N_k^1$, maximum degree $d_{\max}$, validation threshold $\varepsilon_{\rm model}$
\Ensure Local index set $D_k$ and local surrogate $\hat g_k$

\State Set $t\gets 1$
\State Set $E_k^0\gets +\infty$
\Repeat
    \State Define the candidate local index set
    \[
    D_k^t \gets \{\pi_k(1),\ldots,\pi_k(N_k^t)\}
    \]
    \State Split the data indexed by $D_k^t$ into training and test subsets
    \State Set $d\gets 1$
    \State Set $E_{\rm best}\gets +\infty$
    \Repeat
        \State Fit a degree-$d$ local model on the training subset
        \State Compute the test error $E(d)$ on the test subset
        \If{$E(d) < E_{\rm best}$}
            \State Store the current model as the incumbent best model
            \State Set $E_{\rm best}\gets E(d)$
            \State $d\gets d+1$
        \Else
            \State \textbf{break}
        \EndIf
    \Until{$d>d_{\max}$}

    \State Set $E_k^t\gets E_{\rm best}$
    \If{$E_k^t > E_k^{t-1}$}
        \State \textbf{break}
    \Else
        \State Store $D_k^t$ and its incumbent best model
        \State Increase the candidate sample size to obtain $N_k^{t+1}$
        \State $t\gets t+1$
    \EndIf
\Until{stopping condition is met}

\State Set $D_k$ equal to the candidate index set with smallest validation error
\State Set $\hat g_k$ equal to the corresponding fitted local surrogate

\end{algorithmic}
\end{algorithm}

\subsection{Introducing a trust region} 
\label{subsec:tr}
In nonlinear optimization, the concept of a trust region is used to describe a region around the current iterate of an iterative procedure where the local model of the problem built around that iterate is sufficiently close to the original problem.  The trust region is generally defined via a ball for some norm around the current iterate, i.e., $\|x-x_k\| \leq \Delta_k$.  In the conventional setting, the model degree is fixed, and it is known that the local model becomes more and more accurate as the radius of the trust region ball diminishes. 

A related notion appears in the constraint-learning literature, where
data-dependent restrictions, also referred to as \emph{trust regions} or
\emph{validity domains}, are introduced to prevent the optimizer from
exploiting regions that are poorly represented by the data used to construct
the predictive model \cite{opticl,shi2024trust}. This notion differs from the
classical trust-region concept in nonlinear optimization. A classical trust
region is typically centered at the current iterate, and its radius is updated
by comparing the predicted and realized behavior of the problem using new
function evaluations. In the fixed-data constraint-learning setting, no new
evaluations of the unknown constraint can be generated. The region therefore
represents the part of the decision space that is supported by the available
observations.

By the construction of Algorithm~\ref{alg:dcl-local-model}, the prediction accuracy of the local models constructed by DCL cannot be increased (by increasing the complexity of the model, or by increasing the sample size).  We assume that we built the \emph{best} local model possible around the current iterate that is allowed by the distribution of the data.  Therefore, similar to what is done in the work on global constraint learning, we define the trust region via the dataset $D_k$ used to build that model; i.e.  
\[
D_k = \{i: (z_i,y_i) \mbox{ is in training or test set of the learning process at outer iteration } k\}.
\]
In the current implementation of DCL, we consider defining the trust region as the convex hull of the sample points used in $D_k$.  This gives
\begin{equation*}
\label{eq:convex-hull-tr}
\mathcal C_k
=
\left\{
x\in\mathbb{R}^n :
x=\sum_{i\in D_k}\lambda_i z_i,\;
\sum_{i\in D_k}\lambda_i=1,\;
\lambda_i\ge 0
\right\}.
\end{equation*}
With this definition, the subproblem \eqref{eq:subproblem} becomes
\begin{equation*}
\label{eq:dcl-subproblem-hull}
\begin{aligned}
\min_{x,\lambda} \quad & f(x) \\
\text{s.t.}\quad & c(x)\le 0,\\
& \hat g_k(x)\le 0, \\
& x=\sum_{i\in D_k}\lambda_i z_i, \\
& \sum_{i\in D_k}\lambda_i=1, \\
& \lambda_i\ge 0, \qquad i \in D_k, \\
& x\in X.
\end{aligned}
\end{equation*}
This construction ensures that the trial point remains in a region directly supported by the data used to train the local surrogate.






We note that constructions of such a trust region other than the convex hull can be considered.  For instance, Mahalanobis trust region is defined through a data-dependent ellipsoid centered at the current iterate. Let \(\Sigma_k\) denote the covariance matrix of the sample points \(\{z_i:i\in D_k\}\), possibly regularized for numerical stability. Then one may define
\begin{equation*}
\label{eq:mahalanobis-tr}
\mathcal C_k
=
\left\{
x\in\mathbb{R}^n :
(x-x_k)^\top \Sigma_k^\dagger (x-x_k)\le \gamma_k^2
\right\},
\end{equation*}
where \(\Sigma_k^\dagger\) denotes the Moore--Penrose pseudoinverse and \(\gamma_k>0\) is a prescribed radius parameter. 
This trust region adapts its shape to the local spread of the data used in learning.





\bigskip

There is an interesting detail regarding these trust regions.  If the learned model is not feasible within the trust region $\mathcal C_k$, then we have an infeasible subproblem.  That is, subproblem \eqref{eq:subproblem} is feasible if $\mathcal C_k\cap\mathcal F\neq\emptyset$, where $\mathcal F=\{(x_1,x_2): c(x)\le 0, \hat g_k(x)\le 0\}$.  To deal with this situation, we add a penalty term to subproblem \eqref{eq:subproblem} that allows but penalizes the violation of the learned constraint within the trust region as given in \eqref{eq:dcl-relaxed-subproblem}:

\begin{equation}
\label{eq:dcl-relaxed-subproblem}
\begin{aligned}
\min_{x\in X,\ s\ge 0}\quad & f(x)+\rho_k s\\
\text{s.t.}\quad & c(x)\le 0,\\
& \hat g_k(x)\le s,\\
& x\in \mathcal C_k.
\end{aligned}
\end{equation}
The penalty term ensures the subproblem is feasible (provided that the original constraints $c(x)\leq 0$ are consistent in $\mathcal C_k$).

\section{Experimental Results}

We devote this section to our computational study. Starting with a synthetic example illustrating the benefits of using DCL over a global model, we conduct numerical experiments on two case studies from the literature.

\subsection{A Synthetic Example} 
In this subsection, we investigate the potential of DCL in computing better solutions via focused local models.  To be able to fairly assess the solution quality, we work on a synthetic example for which the true constraint function is known.  In particular, we consider minimization of the Rosenbrock function in the following constrained form
\begin{align*}
\min_{x_0,x_1,x_2} \ \ & f(x) = x_0 \\
 & 100(x_2-x_1^2)^2+(1-x_1)^2 \leq x_0.
\end{align*}
The true solution of the problem is $[x_0,x_1,x_2]=[0, 1, 1]$.

\begin{figure}[htbp]
\centering
\begin{minipage}[c]{0.44\textwidth}
  \centering
  \includegraphics[width=\linewidth]{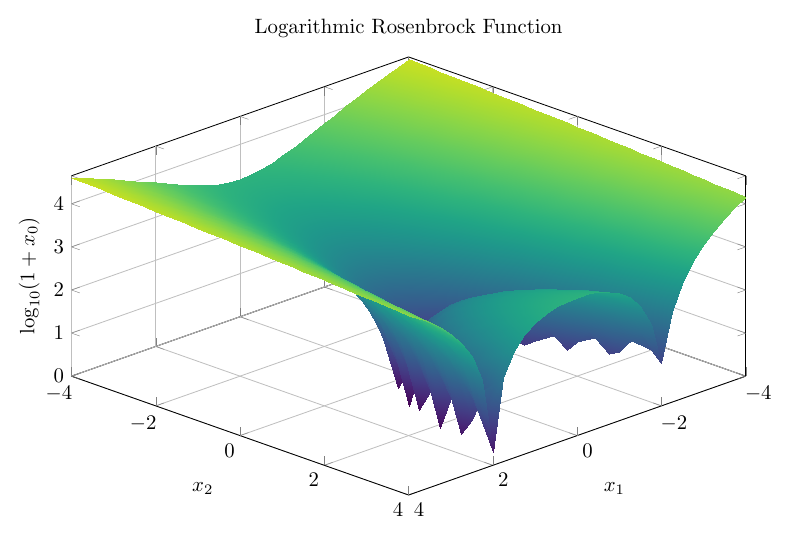}

  \smallskip
  {\small (a)}
\end{minipage}
\hfill
\begin{minipage}[c]{0.10\textwidth}
  \centering

  \pgfplotscolorbardrawstandalone[
    colormap/viridis,
    point meta min=0,
    point meta max=5,
    colorbar style={
      height=4.2cm,
      width=0.30cm,
      ytick={0,1,2,3,4,5},
      ylabel={$\log_{10}(1+x_0)$},
      yticklabel style={font=\small},
      ylabel style={font=\small}
    }
  ]
\end{minipage}
\hfill
\begin{minipage}[c]{0.44\textwidth}
  \centering
  \includegraphics[width=\linewidth]{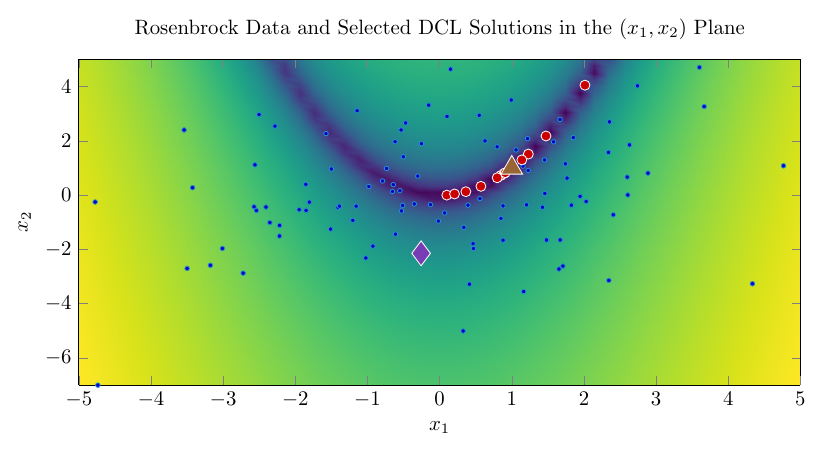}

  \smallskip
  {\small (b)}
\end{minipage}

\caption{Synthetic Rosenbrock example.  
(a) The logarithmically transformed Rosenbrock surface
$\log_{10}(1+x_0)$, where
$x_0=100(x_2-x_1^2)^2+(1-x_1)^2$.
(b) Distribution of the data and selected DCL solutions in the
$(x_1,x_2)$-plane.
The background colors represent the logarithmically scaled Rosenbrock
objective values. Blue circles denote the data points, while red circles
denote the selected DCL iterates with a test relative RMSE below $0.15$.
The brown triangle indicates the true solution, and the purple diamond
indicates the solution obtained using the global surrogate model.
}
\label{fig:rosenbrock_example}
\end{figure}

We pretend as if the constraint is unknown; we generate 100 random data points by adding random noise (from standard normal distribution) to the evaluations obtained via the constraint function $g(x_0,x_1,x_2)=100(x_2-x_1^2)^2+(1-x_1)^2-x_0$.  We assume that only these data points are available to be used in the process of optimization, and there is no way to access an oracle returning $g(x_0,x_1,x_2)$ during the DCL procedure. This setup allows us to observe the operation of DCL on the true problem.  The distribution of the data is shown in Figure~\ref{fig:rosenbrock_example}(b). 

We try two approaches to solve the problem: (i) we use all the data and fit once a global surrogate model, (ii) we do a dynamic local model refit at each iterate as suggested by DCL.  For local model selection and continuation, we use RMSE computed with normalized error terms. 
The same score is used consistently throughout this experiment.

As for the global surrogate, a fifth degree polynomial is chosen in cross validation.  Solution of the  optimization problem with this global model is obtained as
\[
x^{GS}=[-11614.2045, -3.4255, -2.6393].
\]
At this solution point, the true constraint function value is
\[
g(x^{GS})=32292.67,
\]
which is quite far from true feasibility.

\begin{table}[h!]
\caption{DCL iterations for Rosenbrock.}
\label{tab:rosenbrock_updated}
\centering
\scriptsize
\setlength{\tabcolsep}{1pt}
\renewcommand{\arraystretch}{0.95}
\begin{tabular}{c rr
>{\raggedleft\arraybackslash}p{0.06\linewidth} 
>{\raggedleft\arraybackslash}p{0.09\linewidth} 
>{\raggedleft\arraybackslash}p{0.09\linewidth}
>{\raggedleft\arraybackslash}p{0.07\linewidth}
>{\centering\arraybackslash}p{0.07\linewidth}
c 
>{\centering\arraybackslash}p{0.08\linewidth} 
>{\centering\arraybackslash}p{0.08\linewidth}}
\toprule
Iter.
& $x_{0}$
& $x_{1}$
& $x_{2}$
& True $g(x)$
& Global $\hat{g}$
& DCL $\hat{g}$
& Data Size
& Degree
& Train rRMSE
& Test rRMSE \\
\midrule

0
& -1.6372
& 0.2137
& 0.0432
& 2.2561
& 965.5526
& 0.0000
& 54
& 4
& 0.0135
& 0.3375 \\

1
& -0.1152
& 0.7174
& 0.5127
& 0.1955
& 965.5525
& 0.0000
& 66
& 4
& 0.1013
& 0.1020 \\

2
& -4.1169
& 1.8237
& 3.3206
& 4.7985
& 965.5651
& 0.0000
& 54
& 4
& 0.0096
& 0.4326 \\

3
& 0.0471
& 0.8730
& 0.7632
& -0.0308
& 965.5525
& 0.0000
& 78
& 4
& 0.0535
& 0.0446 \\

4
& -70.8254
& 0.0973
& -0.4493
& 92.6930
& 965.6479
& -0.0075
& 30
& 3
& 0.5289
& 0.7394 \\

5
& -0.3342
& 1.1734
& 1.3752
& 0.3646
& 965.5525
& 0.0000
& 78
& 4
& 0.0161
& 0.2142 \\

6
& 1.9978
& -1.5757
& 2.2634
& 9.4567
& 965.5546
& 78.7051
& 30
& 1
& 0.7639
& 0.7090 \\

7
& 1.0147
& -1.4997
& 2.3422
& 6.0994
& 965.5531
& 0.0000
& 54
& 4
& 0.0680
& 0.2711 \\

8
& -0.9786
& 0.8708
& 0.7543
& 0.9969
& 965.5527
& 0.0000
& 78
& 4
& 0.0599
& 0.0183 \\

9
& 33.7810
& 1.4249
& -0.4490
& 581.1819
& 966.0595
& 38.0652
& 30
& 1
& 0.7539
& 0.7760 \\

10
& -53.1788
& 0.2432
& 1.4888
& 258.1488
& 965.8598
& -0.6298
& 30
& 4
& 0.0000
& 0.6957 \\

11
& -0.7951
& 0.8481
& 0.7202
& 0.8183
& 965.5526
& 0.0000
& 78
& 4
& 0.0611
& 0.0495 \\

12
& 2.3431
& -1.5365
& 2.2295
& 5.8164
& 965.5553
& -0.0011
& 30
& 6
& 0.0000
& 0.8269 \\

13
& -0.1749
& 0.8936
& 0.7957
& 0.1870
& 965.5525
& 0.0000
& 78
& 4
& 0.0499
& 0.0838 \\

14
& -4.1369
& 2.0134
& 4.0457
& 5.1708
& 965.5656
& 0.0000
& 66
& 4
& 0.1762
& 0.0875 \\

15
& -0.0835
& 1.4738
& 2.1747
& 0.3087
& 965.5525
& 0.0000
& 66
& 4
& 0.0352
& 0.0947 \\

16
& 33.7810
& 1.4249
& -0.4490
& 581.1818
& 966.0595
& 21.3788
& 30
& 1
& 0.7732
& 0.7095 \\

17
& -0.4830
& 1.2664
& 1.6011
& 0.5547
& 965.5526
& 0.0000
& 66
& 4
& 0.0677
& 0.1539 \\

18
& -0.3457
& 0.7971
& 0.6364
& 0.3869
& 965.5525
& 0.0000
& 78
& 4
& 0.0643
& 0.0727 \\

19
& 1.7708
& 0.0992
& -0.0022
& -0.9450
& 965.5525
& 0.0000
& 66
& 4
& 0.0834
& 0.0572 \\

20
& 0.9071
& 0.2061
& 0.0408
& -0.2766
& 965.5525
& 0.0000
& 54
& 4
& 0.0659
& 0.0779 \\

21
& -39.9506
& 0.3959
& 3.2128
& 974.3044
& 966.1393
& -0.0019
& 30
& 4
& 0.0000
& 0.6776 \\

22
& -1.5439
& 1.1856
& 1.4193
& 1.5969
& 965.5534
& 0.0000
& 54
& 4
& 0.0365
& 0.3441 \\

23
& 0.5460
& 0.5716
& 0.3208
& -0.3591
& 965.5525
& 0.0000
& 66
& 4
& 0.0491
& 0.0691 \\

24
& 0.3145
& 1.2287
& 1.5085
& -0.2621
& 965.5525
& 0.0000
& 78
& 4
& 0.0595
& 0.0377 \\

25
& 1.2601
& 0.3629
& 0.1236
& -0.8476
& 965.5526
& 0.0000
& 66
& 4
& 0.0717
& 0.0788 \\

26
& -0.3254
& 0.9892
& 0.9780
& 0.3255
& 965.5525
& 0.0000
& 78
& 4
& 0.0523
& 0.1151 \\

27
& -0.5381
& 0.9082
& 0.8227
& 0.5470
& 965.5526
& 0.0000
& 78
& 4
& 0.0232
& 0.1266 \\

28
& -0.0433
& 1.1392
& 1.2947
& 0.0637
& 965.5525
& 0.0000
& 78
& 4
& 0.0902
& 0.0472 \\

29
& -16.7957
& 0.3969
& 0.0254
& 18.9065
& 965.5646
& 0.0000
& 30
& 3
& 0.7085
& 0.6316 \\

\bottomrule
\end{tabular}
\end{table}

We let DCL run on the same problem for 30 iterations.  Interestingly, the DCL mechanism chooses to fit fourth degree polynomials at several iterations; however, it uses only subsets of data near the current iterate. The selected local data subset sizes are in the interval $[30,78]$. Lower-degree local models are also selected at some iterations.

In Table~\ref{tab:rosenbrock_updated}, we report a summary of the DCL iterates: the true constraint value $g(x_k)$, the global and local model predictions (global $\hat{g}$ and DCL $\hat{g}$), and information on the local model training (RMSE computed with normalized error terms for train and test sets). 
It is possible to observe that the local DCL model value is close to zero at all iterates.  This is expected since these points are minimizers of the local model and the constraint of the problem is active at that point by the construction of the problem.  There are a few iterations where the iterate is infeasible with respect to the local DCL model so that the relaxation variable (i.e. variable $s$ in subproblem \eqref{eq:dcl-relaxed-subproblem}) takes a positive value.  The true $g(x)$ value is not very far from the local model predictions of DCL as long as the test error is no larger than $0.15$.  

Indeed, it is not natural for DCL to report one of these iterates as the final solution.  We believe that the set of later iterates with small test error form a reliable \emph{region of attraction}.  We observe that the true constraint is satisfied (i.e. $g(x)\leq 0$) in a part of this region.  In Figure~\ref{fig:rosenbrock_example}, we plot the last 20 iterates with relatively small test error (smaller than 0.15 for this instance) to illustrate this region.

\begin{figure}[h]
\centering
\includegraphics[width=1\textwidth]{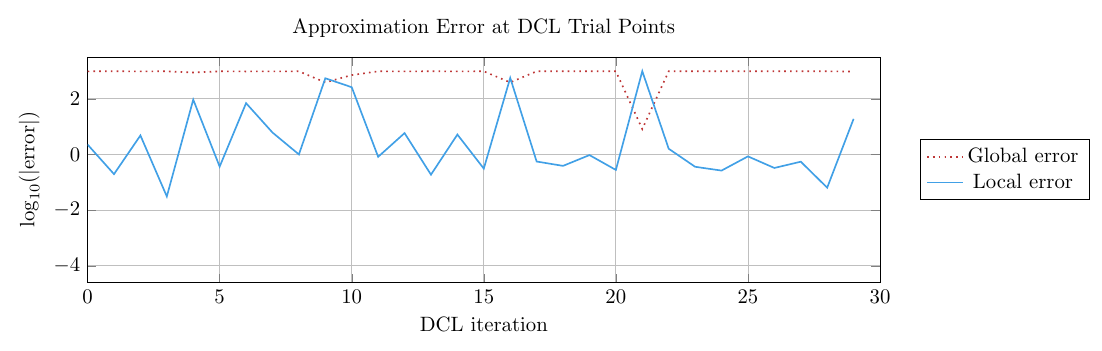} 
\caption{Approximation error comparison of the global model and the local models constructed by DCL.}
\label{fig:rosenbrock_errors} 
\end{figure}

Note also from Table~\ref{tab:rosenbrock_updated} that the global model acts like a bias, while the values of the true constraint function are far from being constant.  In Figure~\ref{fig:rosenbrock_errors}, we plot the error of the global model computed at the iterates of the DCL run, and compare it to that of the local DCL models.  The plotted quantity is
\[
\log_{10}(|\hat g(x_k)-g(x_k)|).
\]
The figure clearly shows that the local DCL models have smaller approximation error at these iterates.


We should mention that the global model fails to be more successful even when we train it with the right hypothesis class (degree 4 polynomial).  This is because the data are uneven and noisy, and with this data the important region (the Rosenbrock valley and its boundary) cannot be captured well with one fit over the whole domain.  
This supports one of the motivations of DCL: local accuracy in the optimization-relevant region can be more useful than a single global fit over the entire dataset.

\bigskip

\subsection{The Palatable Diet Problem} 
This subsection presents an experiment to test our conjecture that using simpler local models and smaller datasets can be computationally more efficient compared to a complex global model using a large amount of data at once.  The test problem is a real-world case studied in \cite{opticl}.  It is a diet problem with a linear objective and linear constraints.  However, there is an additional restriction on \emph{palatability} which does not have a known analytical form.  There is data on palatability values for different food baskets though.  The dataset contains a total of 5000 data points.  The mathematical model of the optimization problem is as follows.  
\begin{align*}
\min_x  \ \ & c^Tx \ \ \ \  \ \ \ \ \ \mbox{(cost of the food basket)}\\
& Ax\geq b \ \ \ \ \ \mbox{(nutritional requirements)}\\
& g(x) \geq b_g \ \  \mbox{(palatability requirement)}\\
& l \leq x \leq u.
\end{align*}
Here, $g(x)$ stands for the unknown function giving the palatability score of a food basket $x$.

In \cite{opticl}, different machine learning models have been tried on this dataset to get an estimator of the palatability score, and the best fit was obtained via a multilayer perceptron.  Therefore, we start by running the DCL code for one iteration with a neural network (NN) model by initializing the size of the dataset as 5000.  Clearly, this is equivalent to a global constraint learning run.  In our choice of the settings for the NN training, we followed the choices of the \cite{opticl} implementation.  We were therefore able to validate our implementation by comparing its output with that of the implementation of \cite{opticl}.  Table~\ref{tab:nn_comparison} gives a summary of that comparison.
\begin{table}[h]
\centering
\caption{Comparison with the implementation of \cite{opticl}}
\label{tab:nn_comparison}
\begin{threeparttable}
\begin{tabular}{
l c c c
}
\toprule
{Model} & {Data Size} & {Train MSE} & {Test MSE} \\
\midrule
Baseline MLP Implementation & 5000 & 0.0110 & 0.0183 \\ 
One-iteration DCL with NN & 5000 & 0.0093 & 0.0107 \\
\bottomrule
\end{tabular}
\begin{tablenotes}[flushleft]
\footnotesize
\item The baseline MLP values are the results reported in \cite{opticl}.
\end{tablenotes}
\end{threeparttable}
\end{table}

Next, we use this global NN surrogate in optimization by stating it with a set of linear constraints and binary variables following again the approach in \cite{opticl}. The overall optimization problem is a mixed integer program, and was solved via Gurobi\cite{gurobi}.  In our implementation, we used Gurobi Machine Learning Package\cite{grbml} to embed the NN model into the mixed integer programming model. The solution for the optimization model with the global NN surrogate is obtained as   
\[
\begin{aligned}
x^{GS} = & (0.1604,\,0,\,0,\,-2.7003\times 10^{-8},\,0,\,0,\,0,\,0.0001,\\
    &0.5678,\,0.05,\,0.3596,\,1.3096,\,0,\,0.0033,\,0,\\
    &0,\,0,\,0,\,0,\,0,\,0.2,\,0.2403,\,2.1521,\,0,\,0.7),
\end{aligned}
\]
with the objective function value 
\[
f(x^{GS}) = 3431.3255.
\]
In Table~\ref{tab:nn_global_results} we provide information on the size of the mixed integer problem solved by Gurobi.  For the NN model, degree=2 corresponds to a single layer with 20 neurons. 

\begin{table}[h]
\centering
\caption{Optimization with the global NN surrogate}
\label{tab:nn_global_results}
\setlength{\tabcolsep}{4pt}
\begin{adjustbox}{max width=\textwidth}
\begin{tabular}{ c c c c c c c c c
}
\toprule
{Iter.} & {Objective} & {Data Size} & {Deg.} & {Train MSE} & {Test MSE} & {MIP Cols} & {MIP Rows} \\
\midrule
1 & 3431.3255 & 5000 & 2 & 0.0093 & 0.0107 & 5091 & 87 \\
\bottomrule
\end{tabular}
\end{adjustbox}
\end{table}

In the second part of the experiment, we try running DCL with local decision tree models which is a smaller complexity model class compared to NN.  After 10 iterations, the solution obtained by DCL is  
\[
\begin{aligned}
x^{DCL} = (&0.1604,\,0,\,0,\,-2.7003\times 10^{-8},\,0,\,0,\,0,\,0.0002,\\
    &0.5678,\,0.05,\,0.3596,\,1.3096,\,0,\,0.0033,\,0,\\
    &0,\,0,\,0,\,0,\,0,\,0.2,\,0.2403,\,2.1521,\,0,\,0.7),
\end{aligned}
\]
with the objective function value 
\[
f(x^{DCL}) = 3431.3255.
\]
This is exactly the same solution as the one found by using the global surrogate.

The information on the DCL iterates are given in Table~\ref{tab:dt_results}.  Note that the train and test errors reported in this table are computed using the subsets of data selected by DCL for model training.  At all iterations, DCL chooses to construct depth four local decision tree models.  The critical observation is that the sizes of the MIP models solved by DCL are significantly smaller than the solve with the global surrogate.  In other words, DCL could compute the same solution for the problem by fitting a sequence of smaller complexity learning models and by solving a sequence of smaller dimensional optimization models.  Moreover, the solution provided by DCL is \emph{explainable} as it is computed based on a depth four decision tree model. 

\begin{table}[h]
\centering
\caption{Optimization with DCL using local decision tree models}
\label{tab:dt_results}
\scriptsize
\begin{threeparttable}
\begin{adjustbox}{max width=\textwidth}
\begin{tabular}{rrrrrrrr}
\toprule
\multicolumn{1}{c}{Iter.} &
\multicolumn{1}{c}{Objective} &
\multicolumn{1}{c}{Data Size} &
\multicolumn{1}{c}{Deg.} &
\multicolumn{1}{c}{Train MSE} &
\multicolumn{1}{c}{Test MSE} &
\multicolumn{1}{c}{MIP Cols} &
\multicolumn{1}{c}{MIP Rows} \\
\midrule
1	&	3451.1221 & 1000 & 4 & 0.0107 & 0.0122 & 1042 & 44 \\
2	&	3431.3255 &  800 & 4 & 0.0111 & 0.0158 &  842 & 44 \\
3  & 3508.9388 &  900 & 4 & 0.0115 & 0.0124 &  942 & 44 \\
4  & 3675.1354 &  700 & 4 & 0.0108 & 0.0135 &  742 & 44 \\
5  & 3540.6080 &  800 & 4 & 0.0124 & 0.0126 &  842 & 44 \\
6  & 3683.5572 &  900 & 4 & 0.0114 & 0.0159 &  942 & 44 \\
7  & 3533.1559 & 1000 & 4 & 0.0132 & 0.0158 & 1042 & 44 \\
8  & 3683.5572 & 1200 & 4 & 0.0132 & 0.0147 & 1242 & 44 \\
9  & 3538.1445 & 1000 & 4 & 0.0134 & 0.0157 & 1042 & 44 \\
10 & 3431.3255 &  600 & 4 & 0.0105 & 0.0117 &  642 & 44 \\
\bottomrule
\end{tabular}
\end{adjustbox}
\end{threeparttable}
\end{table}

\bigskip

\subsection{Concrete Strength Problem}


In this subsection, we apply DCL to a data-driven concrete mixture design problem.
The experiment is based on the concrete compressive-strength case study studied in 
\cite{alcantara2023}.   
Here, each concrete mixture is represented by
\[
x =
(x_{\mathrm{cem}},x_{\mathrm{slag}},x_{\mathrm{fly}},x_{\mathrm{water}},
x_{\mathrm{sup}},x_{\mathrm{coarse}},x_{\mathrm{fine}},d),
\]
where the first seven variables denote material quantities in kg/m$^3$, and
$d$ denotes the age of the concrete in days.  The optimization problem is given as 
\begin{align*}
\min_x \quad & c^\top x \\
\text{s.t.} \quad
& \hat{s}(x) \geq 45, \\
& 2230 \leq
x_{\mathrm{cem}}+x_{\mathrm{slag}}+x_{\mathrm{fly}}+x_{\mathrm{water}}
+x_{\mathrm{sup}}+x_{\mathrm{coarse}}+x_{\mathrm{fine}}
\leq 2450, 
\end{align*}
where $\hat{s}(x)$ denotes the point prediction of the compressive
strength, measured in MPa. Following the reference study, we employ the material cost vector
\[
c =
(0.050,\;0.040,\;0.045,\;0.002,\;1.800,\;0.020,\;0.020,\;0),
\]
where the zero coefficient corresponds to age. This reflects that the objective measures
material production cost.  Age affects the learned strength prediction, but it is not a
material component. The required compressive strength is 45 MPa, and the total
component weight is constrained to lie between 2230 and 2450 kg/m$^3$. 

We consider learning a random-forest global surrogate for point estimation of the compressive strength as done in the reference article.  We try fitting a random forest with 100 estimators, a random forest with 10 estimators, and a CART.  The degree of the models are chosen with cross-validation for each model type.  Table~\ref{tab:global_summary} summarizes the results. A simple CART model elapses less time, but produces a global prediction model with relatively large test error.  The random forest models provide lower test error; thus, a more appropriate choice for a global model.

\begin{table}[h]
\centering
\caption{Results with a global point estimation model}
\label{tab:global_summary}
\scriptsize
\begin{threeparttable}
\begin{adjustbox}{max width=\textwidth}
\begin{tabular}{rrrrrrrrr}
\toprule
Model type & Data size & Degree & Train rRMSE & Test rRMSE & Material obj. & Relax. & MIP time (s) & Total time (s) \\
\midrule
CART & 1030 & 11 & 0.0595 & 0.1755 & 49.1095 & 0.0000 & 0.2990 & 1.9806 \\
RF (10 Est) & 1030 & 16 & 0.0776 & 0.1557 & 49.0494 & 0.0000 & 7.4865 & 9.9771\\
RF (100 Est) & 1030 & 16 & 0.0660 & 0.1502 & 52.9186 & 0.0000 & 119.4762 & 138.3061 \\
\bottomrule
\end{tabular}
\end{adjustbox}
\end{threeparttable}
\end{table}



As for the DCL run, CART models are considered.  The maximum tree depth is treated as the model degree, and candidate depths from 1 to 20 are considered.  The output of the DCL run is summarized in Table~\ref{tab:dcl_iteration_summary_cart}.  In the last column of the table, the cumulative total time value is reported. Local CART models can achieve test error values comparable or lower than the values for global RF models reported in Table~\ref{tab:global_summary}.  At some iterates, it is able to achieve lower objective values compared to the ones provided by the global surrogates, with local prediction models having small test errors. 

\begin{table}[h]
\centering
\caption{DCL iteration summary for the concrete instance: CART}
\label{tab:dcl_iteration_summary_cart}
\scriptsize
\begin{threeparttable}
\begin{adjustbox}{max width=\textwidth}
\begin{tabular}{rrrrrrrrr}
\toprule
Iter. & Data size & Degree & Train rRMSE & Test rRMSE & Material obj. & Relax. & MIP time (s) & Total time (cum.) \\
\midrule
0	&	50	&	8	&	0.0000	&	0.1315	&	51.1408	&	0.0000	&	0.0402	&	0.6732	\\
1	&	82	&	3	&	0.1613	&	0.1995	&	62.4163	&	0.0000	&	0.0752	&	1.7541	\\
2	&	50	&	3	&	0.1517	&	0.2386	&	49.6339	&	4.7319	&	0.0216	&	2.5022	\\
3	&	82	&	6	&	0.0406	&	0.1504	&	51.1458	&	0.0000	&	0.0413	&	3.5768	\\
4	&	82	&	7	&	0.0342	&	0.1512	&	49.7359	&	0.0000	&	0.0396	&	4.6470	\\
5	&	50	&	3	&	0.1412	&	0.1561	&	49.3275	&	0.0000	&	0.0246	&	5.3070	\\
6	&	50	&	6	&	0.0063	&	0.1485	&	48.8290	&	0.0000	&	0.0358	&	6.0094	\\
7	&	82	&	8	&	0.0038	&	0.1355	&	48.9142	&	0.0000	&	0.0539	&	7.1564	\\
8	&	82	&	3	&	0.1543	&	0.1346	&	48.7805	&	0.0000	&	0.0204	&	7.9773	\\
9	&	50	&	3	&	0.1144	&	0.1434	&	48.6100	&	0.2216	&	0.0310	&	8.6675	\\
10	&	82	&	5	&	0.0555	&	0.1797	&	52.0125	&	0.0000	&	0.0389	&	9.6328	\\
11	&	50	&	7	&	0.0004	&	0.1546	&	51.4042	&	0.0000	&	0.0293	&	10.2076	\\
12	&	50	&	9	&	0.0000	&	0.1417	&	50.0223	&	0.0000	&	0.0265	&	10.7263	\\
13	&	50	&	7	&	0.0070	&	0.1513	&	49.6320	&	0.0000	&	0.0248	&	11.2812	\\
14	&	82	&	8	&	0.0146	&	0.1220	&	49.6331	&	0.0000	&	0.0412	&	12.1955	\\
15	&	50	&	3	&	0.1318	&	0.1971	&	49.6124	&	0.2694	&	0.0179	&	12.8267	\\
16	&	146	&	10	&	0.0007	&	0.1992	&	49.3522	&	0.0000	&	0.0545	&	14.4570	\\
17	&	114	&	10	&	0.0007	&	0.1905	&	48.7384	&	0.0000	&	0.0521	&	15.7335	\\
18	&	82	&	12	&	0.0001	&	0.1843	&	49.8629	&	0.0000	&	0.0422	&	16.8020	\\
19	&	50	&	8	&	0.0000	&	0.1855	&	49.9847	&	0.0000	&	0.0398	&	17.3428	\\
20	&	146	&	14	&	0.0000	&	0.1452	&	49.3275	&	0.0000	&	0.0679	&	18.6664	\\
21	&	82	&	10	&	0.0000	&	0.1475	&	49.9399	&	0.0000	&	0.0422	&	19.6012	\\
22	&	82	&	11	&	0.0000	&	0.1342	&	49.0888	&	0.0000	&	0.0641	&	20.7091	\\
23	&	82	&	3	&	0.1577	&	0.1640	&	48.7000	&	0.0000	&	0.0194	&	21.6210	\\
24	&	82	&	5	&	0.0525	&	0.1810	&	52.0125	&	0.0000	&	0.0235	&	22.3782	\\
25	&	82	&	11	&	0.0000	&	0.1528	&	49.8771	&	0.0000	&	0.0436	&	23.1674	\\
26	&	82	&	6	&	0.0306	&	0.1330	&	49.3275	&	0.0000	&	0.0334	&	23.9505	\\
27	&	50	&	8	&	0.0000	&	0.1603	&	49.0882	&	0.0000	&	0.0429	&	24.5259	\\
28	&	50	&	5	&	0.0458	&	0.1465	&	48.5877	&	2.9318	&	0.0290	&	25.1675	\\
29	&	146	&	7	&	0.0528	&	0.1802	&	49.2904	&	0.0000	&	0.0781	&	27.1241	\\
30	&	82	&	3	&	0.1573	&	0.1736	&	48.5624	&	0.0000	&	0.0225	&	28.2830	\\
31	&	50	&	3	&	0.1169	&	0.2108	&	48.1039	&	5.0507	&	0.0218	&	28.8684	\\
32	&	146	&	13	&	0.0000	&	0.1782	&	49.4297	&	0.0000	&	0.0768	&	30.5579	\\
33	&	50	&	9	&	0.0002	&	0.1454	&	49.8606	&	1.7491	&	0.0340	&	31.3353	\\
34	&	82	&	6	&	0.0418	&	0.1856	&	50.9602	&	0.0000	&	0.0248	&	32.1823	\\
35	&	50	&	3	&	0.1175	&	0.2195	&	50.4641	&	5.7780	&	0.0233	&	32.6315	\\
36	&	82	&	6	&	0.0445	&	0.1657	&	51.1458	&	0.0000	&	0.0315	&	33.5386	\\
37	&	82	&	4	&	0.0836	&	0.2096	&	49.8715	&	0.0000	&	0.0163	&	34.1729	\\
38	&	50	&	3	&	0.1143	&	0.1625	&	49.6277	&	0.0000	&	0.0237	&	34.7315	\\
39	&	82	&	9	&	0.0003	&	0.1347	&	49.3275	&	0.0000	&	0.0314	&	35.6392	\\
40	&	50	&	6	&	0.0104	&	0.1810	&	50.9097	&	1.4347	&	0.0177	&	35.9909	\\
41	&	114	&	11	&	0.0000	&	0.1776	&	50.0507	&	0.0000	&	0.0729	&	37.2473	\\
42	&	50	&	9	&	0.0000	&	0.1365	&	49.6277	&	0.0000	&	0.0353	&	37.9437	\\
43	&	50	&	3	&	0.1002	&	0.1401	&	50.1820	&	0.0000	&	0.0242	&	38.6372	\\
44	&	50	&	10	&	0.0037	&	0.1290	&	49.8769	&	0.0000	&	0.0310	&	39.3403	\\
45	&	82	&	10	&	0.0000	&	0.1359	&	50.0593	&	0.0000	&	0.0464	&	40.3999	\\
46	&	50	&	3	&	0.1208	&	0.1698	&	49.6277	&	0.0000	&	0.0184	&	40.9170	\\
47	&	82	&	4	&	0.1335	&	0.1922	&	49.7186	&	0.0000	&	0.0302	&	41.8456	\\
48	&	50	&	6	&	0.0114	&	0.1162	&	49.1228	&	1.7491	&	0.0290	&	42.4071	\\
49	&	50	&	3	&	0.0976	&	0.2506	&	49.2981	&	4.0869	&	0.0212	&	43.1949	\\
\bottomrule
\end{tabular}
\end{adjustbox}
\end{threeparttable}
\end{table}




In this experiment, we would like to compare the cost of producing a region of attraction with DCL to the cost of producing a point estimate with a global model.  The smallest achievable test error depends on the data.  For a global model, the best test error is obtained via a relatively complex model class (RF with 100 estimators), which has relatively large computational cost both for training and MIP solve.  As DCL works with local models, it is able to generate models with comparable test errors via the simpler CART class.  These simpler models require significantly less time so that within the time elapsed by the run with a global model, DCL produces a region of attraction (several iterates within a set with small test errors) providing solutions no worse than the one computed via the global model.  



\section{Discussion}
\label{sec:discussion}
The experimental results support the central conjecture of DCL: local surrogates fitted to data in the neighborhood of each iterate can match or improve upon the solution quality of a single global surrogate, while using substantially simpler model classes and solving smaller optimization subproblems. 
Several discussion points remain, and we present them in turn.

\paragraph{Solution quality assessment.} Assessing the quality of the solution returned by DCL is non-trivial in general. In the synthetic Rosenbrock experiment we could evaluate the true constraint function at the returned point; in the diet experiment the global and local approaches returned the same solution, so comparison was straightforward. In the typical setting, however, neither the true constraint function nor a ground-truth optimal solution is available. The challenge is most acute when the learned constraint is (nearly) active at the final point because even a small surrogate error can then determine whether that point is feasible with respect to the true constraint. A principled treatment would require quantifying the uncertainty of the learned surrogate (for instance through conformal prediction intervals or Bayesian credible sets) and propagating that uncertainty into a feasibility certificate. The trade-off between a lower objective value, feasibility with respect to the analytic constraints, and robustness with respect to the noisy learned constraint is ultimately a modeling choice. We plan to study systematic approaches for making this trade-off explicit in future work.


\paragraph{Termination criteria.} The current termination rule (stop when the computed step length falls below a threshold $\varepsilon_{\text{step}}$) proved difficult to trigger in our experiments because the iterates tend to oscillate in a neighborhood of the boundary of the learned feasible region rather than converging monotonically. This behavior is partly an artifact of the stochastic nature of DCL: changing local data neighborhoods and varying validation splits produce different surrogates at nearby iterates. A more robust stopping criterion should be tied to the quality of the current solution rather than the norm of the step. Natural candidates include (i) a bound on the relative change in objective value over a window of iterations, (ii) a statistical test on whether the local surrogate error has fallen below a prescribed tolerance, or (iii) a feasibility check using a held-out validation set.
Designing a termination criterion that is both computationally cheap and theoretically grounded is an important direction for future work.


\paragraph{Convergence analysis.} DCL fits naturally into the stochastic approximation framework, since the local surrogate at each iteration is a noisy estimate of the true constraint function. Unlike the classical SA setting, however, the noise distribution in DCL is iteration-dependent: it depends on the local data subset $\mathcal{D}_k$, the model class selected at iteration $k$, and the realization of the train/test split. The data-supported trust region $\mathcal{C}_k$ introduces an additional complication absent in standard SA theory, because the feasible set of the subproblem itself changes with $k$. A rigorous convergence theory for DCL would need to characterize conditions under which the sequence of iterates approaches a stationary point or a feasible point of the true problem, accounting for both the approximation error of the learned surrogate and the geometry of the data-supported trust region. Connections to model-based DFO convergence theory \cite{conn2009} and to online learning with constraints may provide useful starting points.


\paragraph{Alternative trust region constructions.}
The current implementation defines the trust region $\mathcal{C}_k$ as the convex hull of the local data subset $\mathcal{D}_k$. While this construction has a natural statistical interpretation---the optimizer is not allowed to extrapolate beyond the support of the training data---it introduces a convex-combination constraint that inflates the subproblem and may be overly conservative when the data are irregularly distributed. The Mahalanobis ellipsoid discussed in Section \ref{subsec:tr} is a computationally lighter alternative that adapts to the local covariance structure of the data. Other possibilities include $\ell_p$-balls centered at $x_k$ with radius chosen to enclose a prescribed fraction of $\mathcal{D}_k$, or data-depth regions such as Tukey halfspace depths. A systematic comparison of trust region geometries, both in terms of computational cost and optimization performance, is a natural next step.

\paragraph{Ensemble of local DCL models as a learned new global predictor.} The sequence of local surrogates produced by DCL along the optimization path constitutes a structured collection of models, each accurate in one neighborhood of the domain. The idea of aggregating such local models into a global predictor has recently been studied under the name \emph{subset stacking} \cite{birbil26}, where it was shown that ensembles of models trained on overlapping subsets can recover global prediction accuracy with lower computational cost than a single global fit. An intriguing open question is whether the DCL-generated ensemble carries additional predictive value, and whether this ensemble can be certified as a valid global surrogate for downstream use.


\paragraph{Locally weighted regression.}
The local surrogate in DCL is currently fitted by ordinary (unweighted) regression on the nearest-neighbor subset $\mathcal{D}_k$. A more principled alternative is locally weighted regression \cite{hastie2009}, in which every data point in the full dataset $\mathcal{D}$ receives a weight that decays with its distance from the current iterate $x_k$. This avoids the binary inclusion/exclusion decision implicit in choosing a subset radius, allows for a smooth transition between local and global models, and is well suited to settings where data density varies across the domain. The main open question is how to integrate the resulting weighted surrogate with a data-supported trust region, since the notion of a ``support region'' for a weighted model is less well-defined than for a nearest-neighbor subset.


\section{Conclusion}
\label{sec:conclusion}

We have introduced Dynamic Constraint Learning (DCL), a data-driven framework for constrained optimization in which the analytic form of at least one constraint function is unknown and unqueryable. Rather than fitting a single global surrogate to all available data before optimization begins, DCL integrates learning and optimization: at each iteration it fits a local surrogate to the data nearest the current iterate, solves a subproblem confined to a data-supported trust region, and updates the iterate accordingly.

The key insights motivating DCL are (i) that optimization visits only a low-dimensional path through the decision space, so a surrogate that is accurate near each iterate is more valuable than one that is accurate everywhere, (ii) that local accuracy can be achieved with simpler model classes, which in turn yield easier subproblems, and (iii) that restricting the step to the convex hull of the local training data provides a natural, data-adaptive trust region that keeps the surrogate in its regime of reliability.

Our computational study confirms these insights on one synthetic test problem and two case studies from the literature. On the synthetic instance, DCL finds a feasible point with near-zero true constraint violation, whereas the global surrogate fitted with a larger dataset returns a highly infeasible point due to the uneven distribution of the training data. On the first case study (palatable diet problem) DCL matches the optimal solution found by the global neural-network approach while constructing simpler models and solving a sequence of substantially smaller MIP subproblems, achieving the same solution quality with significantly lower per-iteration computational cost. In the second case study (concrete strength problem), it is observed that DCL can produce a sample set of solutions of comparable quality within the time required by a global surrogate for computing a single solution point.

Several theoretical and practical questions remain open. These are principled termination criteria, systematic comparison of trust region geometries, and convergence analysis.  We view these as natural next steps toward a complete algorithmic and theoretical treatment of DCL. Beyond the immediate extensions, DCL opens a broader research agenda at the interface of machine learning, stochastic approximation, and constrained optimization: it demonstrates that \emph{when} and \emph{where} a surrogate is learned matters as much as \emph{what} is learned, and that tightly coupling the learning and optimization phases can yield both computational and solution-quality benefits that neither phase achieves alone.

\bigskip

\bibliographystyle{alpha}
\bibliography{NLPwDCL}

\end{document}